\documentclass[11pt]{article}
\pdfoutput=1

\usepackage{geometry}
\usepackage{graphicx}
\usepackage{amssymb}
\usepackage{amsmath}
\usepackage{epstopdf}
\usepackage{hyperref}
\usepackage{booktabs}
\usepackage{nth}
\usepackage{url}
\usepackage{float}
\usepackage{tikz}
\usepackage{pgfplots}
\usepackage{siunitx}
\usepackage{multirow}
\usepackage{mathptmx}      
\usepackage{latexsym}
\hypersetup{
    colorlinks=true,
    linkcolor=black,
    citecolor=black,
    filecolor=black,
    urlcolor=black
}

\DeclareMathOperator*{\simx}{sim}

\usepackage{acl2015}
\usepackage{times}
\usepackage{url}
\usepackage{latexsym}

\usepackage[round]{natbib}

\title{Evaluating Unsupervised Dutch Word Embeddings \\ as a Linguistic Resource}

\author{St\'{e}phan Tulkens \and Chris Emmery \and Walter Daelemans \\
  CLiPS, University of Antwerp \\
  {\tt \{first.lastname\}@uantwerpen.be} \\}

\begin{document}
\maketitle

\begin{abstract}
Word embeddings have recently seen a strong increase in interest as a result of
strong performance gains on a variety of tasks. However, most of this research
also underlined the importance of benchmark datasets, and the difficulty of
constructing these for a variety of language-specific tasks. Still, many of the
datasets used in these tasks could prove to be fruitful linguistic resources,
allowing for unique observations into language use and variability. In this
paper we demonstrate the performance of multiple types of embeddings, created
with both count and prediction-based architectures on a variety of corpora, in
two language-specific tasks: relation evaluation, and dialect identification.
For the latter, we compare unsupervised methods with a traditional, hand-crafted
dictionary. With this research, we provide the embeddings themselves, the
relation evaluation task benchmark for use in further research, and demonstrate
how the benchmarked embeddings prove a useful unsupervised linguistic resource,
effectively used in a downstream task.
\end{abstract}

\section{Introduction}

The strong variability of language use within, and across textual media
\citep{collins1977,linell1982} has on many occasions been marked as an important
challenge for research in the area of computational linguistics
\citep{resnik1999,rosenfeld2000}, in particular in applications to social media
\citep{gouws2011}. Formal and informal varieties, as well as an abundance of
deviations from grammar and spelling conventions in the latter, drastically
complicate computationally interpreting the meaning of, and relations between
words. This task of understanding lies at the heart of natural language
processing (NLP). Neural-network-based language models such as the models in
\texttt{word2vec} have recently gained strong interest in NLP due to the fact
that they improved state-of-the-art performance on a variety of tasks in the
field. Given these developments, we found it surprising that only one set of
word embeddings has been publicly released for Dutch \citep{polygot}, which does
not offer sufficiently large dimensionality for state-of-the-art performance.
The primary goal of this research is thus evaluating word embeddings derived
from several popular Dutch corpora and the impact of these sources on their
quality, specifically focusing on problems characteristic for Dutch. Word
embeddings---being an unsupervised technique---cannot be easily evaluated
without comparing performance in some downstream task. Therefore, we present
two novel benchmarking tasks of our own making: a relation identification task
analogous to previous evaluations on English, in which the quality of different
kinds of word embeddings is measured, and a dialect identification task which
measures the usefulness of word embeddings as a linguistic resource for Dutch
in particular. In the literature, there has been some debate on the
effectiveness of prediction-based embeddings when compared to more classical
count-based embedding models \citep{baroni2014}. As such, we train both count-
(SPPMI) and prediction-based (SGNS) models, and compare them to previous efforts
in both Dutch and English. Additionally, we make the trained embeddings, the
means to construct these models on new corpora, as well as the materials to
evaluate their quality available to the research community\footnote{Code and
data are accessible via \url{github.com/clips/dutchembeddings}.}.

\section{Related Work}

An idea mostly brought forward by the earlier distributional semantic models
(DSMs), is that the context in which words occur (the distribution of the words
surrounding them) can serve as a representation of their meaning, also known as
the distributional hypothesis \citep{harris1954}. Count based DSMs include LSA
\citep{deerwester1989,deerwester1990}, PLSA \citep{hofmann1999} and LDA
\citep{blei2003latent}, which first create an explicit matrix of occurrence
counts for a number of documents, and then factor this matrix into a
low-dimensional, dense representation using Singular Value Decomposition (SVD)
\citep{schutze1997projections}. A more explicit way of implementing the
distributional hypothesis is through the use of matrices containing
co-occurrence counts \citep{lund1996}, which are then optionally transformed
through the use of some information-theoretic measure, such as PMI (Pointwise
Mutual Information) \citep{bullinaria2007,levy2014neural} or entropy
\citep{rohde2006}. Over the years, these DSMs have proven adequate as a semantic
representation in a variety of NLP tasks.

An alternative to these count-based methods can be found in models predicting
word identity from a given sentence context. Rather than deriving meaning from
the representation of an entire corpus, these construct word representations one
sentence at a time. In attempting to predict the current word through its
context, the model will learn that words which occur in similar sentence
contexts are semantically related. These representations are projected into
$n$-dimensional vector spaces in which more similar words are closer together,
and are therefore referred to as \emph{word embeddings}. Recently, several
models which create prediction-based word embeddings
\citep{bengio2006,collobert2008,mnih2009,mikolov2013b,pennington2014} have proved
successful \citep{turian2010,collobert2011,baroni2014} and consequently have
quickly found their way into many applications of NLP. Following
\cite{goldberg2014}, we call the embeddings represented by dense vectors
\emph{implicit}, as it is not immediately clear what each dimension represents.
Matrix-based sparse embeddings are then called \emph{explicit} as each dimension
represents a separate context, which is more easily interpretable. One of the
more successful and most popular methods for creating word embeddings is
\texttt{word2vec} \citep{mikolov2013,mikolov2013b}. While \texttt{word2vec} often
referred to as a single model, it is actually a collection of two different
architectures, SkipGram (SG) and Continuous Bag of Words (CBoW), and two
different training methods, hierarchical skipgram (HS) and negative sampling
(NS). \cite{levy2015improving} show that one of the architectures in the
\texttt{word2vec} toolkit, SkipGram with Negative Sampling (SGNS) implicitly
factorizes a co-occurrence matrix which has been shifted by a factor of
$\log(k)$, where $k$ is the number of negative samples. Negative samples, in
this case, are noise words which do not belong to the context currently being
modelled. Subsequently, the authors propose SPPMI, which is the explicit,
count-based version of SGNS, i.e. it explicitly creates a co-occurrence matrix,
and then shifts all cells in the matrix by $\log(k)$. SPPMI is therefore a
count-based model which is theoretically equivalent to SGNS. When compared to
other methods, such as GloVe \citep{pennington2014}, SPPMI has showed increased
performance \citep{levy2015improving}.

\section{Data}

In our research, we used four large corpora, as well as a combination of three
of these corpora to train both SPPMI and \texttt{word2vec}. Additionally, we
retrieved a dataset of region-labeled Dutch social media posts, as well as
hand-crafted dictionaries for the dialect identification task (see
\ref{dialect}).

\subsection{Corpora}\label{corpora}

\paragraph{Roularta} The Roularta corpus \citep{roularta} was compiled from a set
of articles from the Belgian publishing consortium
Roularta\footnote{\url{www.roularta.be/en}}. Hence, the articles in this
corpus display more characteristics of formal language than the other corpora.

\paragraph{Wikipedia} We created a corpus of a Wikipedia dump\footnote{The
\texttt{2015.07.03} dump, available at:
\url{dumps.wikimedia.org/nlwiki/20150703}, retrieved on the 29/06/2015.}.
The raw dump was then parsed using a Wikipedia parser,
\texttt{wikiextractor}\footnote{\url{github.com/attardi/wikiextractor}},
and tokenized using \texttt{Pattern} \citep{patternref}.

\paragraph{SoNaR} The SoNaR corpus \citep{sonar500} is compiled from a large
number of disparate sources, including newsletters, press releases, books,
magazines and newspapers. It therefore displays a
high amount of variance in terms of word use and style. Unlike the COW corpus
(see below), some spelling variation in the SoNaR corpus is automatically
corrected, and the frequency of other languages in the corpus is reduced through
the use of computational methods.

\paragraph{COW} The COW corpus \citep{COW} is a ~4 billion word corpus which was
automatically retrieved from domains from the .be and .nl top level domains in
2011 and 2014 \citep{COW}. As such, there is considerable language variability in
the corpus. The corpus was automatically tokenized, although we did perform some
extra pre-processing (see \ref{preprocessing}).

\paragraph{Social Media Dataset} The social media dataset was retrieved from
several Dutch Facebook pages which all had the peculiarities of a specific
dialect or province as their subject. As such, these pages contain a high
percentage of dialect language utterances specific to that province or city. For
each of these Facebook pages, the region of the page was determined, and all
posts on these pages were then labelled as belonging to this region, resulting
in a corpus of ~96,000 posts. Tokenization and lemmatization of each post was
performed using \texttt{Frog} \citep{bosch2007}. This dataset is noisy in nature,
and weakly labelled, as people might use standard language when talking about
their province or home town, or will not use the `correct' dialect on the
designated page. This will prove the robustness of our models, and specifically
that of our methods for ranking dialects.

\paragraph{Combined} In addition to these corpora, we also created a Combined
corpus, which consists of the concatenation of the Roularta, Wikipedia and SoNaR
corpora, as described above. We created the Combined corpus to test whether
adding more data would improve performance, and to observe whether the pattern
of performance on our relation task would change as a result of the
concatenation.
\begin{table}
\centering
\small
\begin{tabular*}{\columnwidth}{@{\extracolsep{\fill}} lrrrrr}
\toprule
         & \textbf{Roul} & \textbf{Wiki} & \textbf{SoNaR} & \textbf{Comb} & \textbf{COW} \\
\midrule
\textbf{\# S}  &  1.7m & 24.8m  & 28.1m   & 54.8m   &  251.8m  \\
\textbf{\# W}  & 27.7m & 392.0m & 392.8m  & 803.0m  &    4b  \\
\bottomrule
\end{tabular*}
\caption{Sentence and word frequencies for the Roularta, Wikipedia, SoNaR500, Combined and COW corpora, where `m' is million and `b' billion.}
\label{corpus}
\end{table}

\subsection{Preprocessing} \label{preprocessing}

Given that all corpora were already tokenized, all tokens were lowercased, and
those solely consisting of non-alphanumeric characters were excluded.
Furthermore, sentences that were shorter than five tokens were removed, as these
do not contain enough context words to provide meaningful results. Some
additional preprocessing was performed on the COW corpus: as a side-effect of
adapting the already tokenized version of the corpus, the Dutch section contains
some incorrectly tokenized plurals, e.g. \texttt{regio's}, tokenized as
\texttt{regi + o + ' + s}. Given this, we chose to remove all tokens that only
consisted of one character, except the token \texttt{u}, which is a Dutch
pronoun indicating politeness.

\subsection{Dictionaries} \label{sec:dd}

\begin{table}[t]
\centering
\small
\begin{tabular*}{\columnwidth}{@{\extracolsep{\fill}} llrr}
\toprule
\textbf{Province} & \textbf{ID} & \textbf{\# words dict} 	& \textbf{\# posts test}\\
\midrule
Antwerpen         & ANT         & 10,108					& 	20,340 \\
Drenthe           & -           & 1,308						&	0 \\
Flevoland         & -           & 1,794						&	0 \\
Friesland         & FRI         & 4,010						&	1,666 \\
Gelderland        & GEL         & 10,313					&	6,743 \\
Groningen         & GRO         & 7,843						&	147 \\
Limburg           & LI          & 45,337					&	10,259 \\
Noord-Brabant     & N-BR        & 20,380					&	1,979 \\
Noord-Holland     & N-HO        & 6,497						&	2,297 \\
Oost-Vlaanderen   & O-VL        & 23,947					&	14,494 \\
Overijssel        & -           & 4,138						&	0 \\
Utrecht           & UTR         & 1,130						&	7,672 \\
Vlaams-Brabant    & VL-BR       & 7,040						&	5,638 \\
West-Vlaanderen   & W-VL        & 16,031					&	12,344 \\
Zeeland           & ZEE         & 4,260						&	1,562 \\
Zuid-Holland      & Z-HO        & 6,374 					&	11,221 \\
\midrule
Standard Dutch    &             & 133,768					& -\\
\bottomrule
\end{tabular*}
\caption{The type frequencies of the dialect dictionaries, the ID used in Figure 1, the type frequency of the corresponding dictionary, and the number of posts for that province in the test set.}
\label{dialectfreq}
\end{table}

To compare our embeddings to a hand-crafted linguistic resource, we collected a
dictionary containing dialect words and sentences, as well as one for standard
Dutch. The dialect dictionary was retrieved from MWB (Mijn
WoordenBoek)\footnote{\url{www.mijnwoordenboek.nl/dialecten}, retrieved
on 05/10/2015.}, which offers user-submitted dialect words, sentences and
sayings, and their translations. Only the dialect part was retained, and split
in single words, which were then stored according to the region it was assigned
to by MWB, and the province this region is part of. Any overlapping words across
dialects were removed. As a reference dictionary for standard Dutch, the
OpenTaal word list\footnote{Retrieved from
\url{www.opentaal.org/bestanden} on 19/10/2015, version dated 24/08/2011.}
was used. Additionally, it was used to remove any general words from the dialect
dictionaries, i.e. if a word occurred in both the Dutch reference dictionary and
a dialect, it was deleted from the dialect. While employing hand-crafted
dictionaries can be beneficial in many tasks, producing such resources is
expensive, and often takes expert knowledge. Techniques able to use unlabelled
data would not only avoid this, but could also prove to be more effective.

\section{Experiments}

For the evaluation of our Dutch word embeddings, we constructed both a novel
benchmark task and downstream task, which can be used to evaluate the
performance of new embeddings for Dutch.

\subsection{Parameter Estimation}

For each corpus, we trained models using the \texttt{word2vec} implementation
\citep{rehurek2010,mikolov2013} from
\texttt{gensim}\footnote{\url{radimrehurek.com/gensim/}}. In order to
determine optimal settings for the hyperparameters, several models were trained
with different parameter values in parallel and were evaluated in the relation
evaluation task (see below). For \texttt{word2vec} the SGNS architecture with a
negative sampling of 15, a vector size of 320, and a window size of 11 maximized
the quality across all corpora. For the SPPMI models, we created embeddings for
the 50,000 most frequent words, experimenting with window sizes of 5 and 10, and
shift constants of 1, 5 and 10. For all models, a shift constant of 1 and a
window size of 5 produced the best results, the exception being the model based
on the Roularta corpus, which performed best with a shift constant of 5 and a
window size of 5. Relying on only one set of hyperparameters, as well as the
performance of the relation task, could be seen as a point of contention.
However, we argue in line with \cite{schnabel2015} that `true' performance
across unrelated downstream tasks is complicated to assess. Nevertheless, we
regard our approach to be satisfactory for the research presented here. Finally,
in addition to our own models, we use the Polyglot embeddings \citep{polygot} as
a baseline, as this is currently the only available set of embeddings for Dutch.

\subsection{Relation Identification}

\begin{table}
\centering
\small
\begin{tabular}{lll}
\toprule
              &   \textbf{Example} & \textbf{Translation}  \\
\midrule
\textbf{Superlative}   & `slecht' - `slechtst'        & bad - worst           \\
\textbf{Past Tense}    & `loop' - `liep' & walk - walked   \\
\textbf{Infinitive}    & `dans' - `dansen'            & dance - danced        \\
\textbf{Comparative}   & `groot' - `groter'           & big - bigger          \\
\textbf{Diminutive}    & `koe'- `koetje'      & cow - small cow \\
\textbf{Plural}        & `boek' - `boeken'     & book - books  						\\
\textbf{Opposites}     & `mooi' - `lelijk'  & beautiful - ugly         \\
\textbf{Currency}      & `japan' - `yen'              &                       \\
\textbf{Nationalities} & `belgi\"{e}' - `belg'        & belgium - belgian     \\
\textbf{Country}       & `noorwegen' - `oslo'         & norway - oslo         \\
\textbf{Gender}        & `oom' - `tante'              & uncle - aunt          \\
\bottomrule
\end{tabular}
\caption{Relation Evaluation set categories, examples, and translation of examples.}
\label{predtable}
\end{table}
%

\begin{table*}
\centering
\scalebox{0.75}{
\begin{tabular}{lrrrrrrrrrrrrrrrrr}
\toprule
\multirow{1}{*}{} &
	 & &
	\multicolumn{2}{c}{\textbf{Wikipedia}} & &
	\multicolumn{2}{c}{\textbf{Roularta}} & &
	\multicolumn{2}{c}{\textbf{SoNaR500}} & &
	\multicolumn{2}{c}{\textbf{Combined}} & &
	\multicolumn{2}{c}{\textbf{COW}} \\

						 & \textbf{Polyglot}     & & {SPPMI} & {SGNS} & & {SPPMI} & {SGNS}	& & {SPPMI} & {SGNS} 		& & {SPPMI} & {SGNS}	 		 & & {SPPMI} 		& {SGNS}	\\

\midrule
\textbf{Superlative} 	 & 13.3 & & 0.6     & 8.3	& & 0.3     & 10.0  & & 0.0  	& 15.5		  	& & 0.0  	& 15.1		     & & 0				& \textbf{39.9}\\
\textbf{Past Tense} 	 & 5.8 	& & 20.7    & 37.8  & & 16.4    & 41.2  & & 25.8 	& \textbf{68.3} & & 26.9 	& 46.2		     & & 25				& 66.1\\
\textbf{Infinitive} 	 & 1.8 	& & 12.1    & 14.0  & & 7.7     & 19.0  & & 41.2 	& 63.1		  	& & 36.2 	& 18.0		     & & 59 			& \textbf{65.0}\\
\textbf{Comparative} 	 & 18.6 & & 12.1    & 39.0  & & 17.6    & 43.8  & & 41.2 	& 63.4		  	& & 40.0 	& 55.5		     & & 53.7 			& \textbf{76.6}\\
\textbf{Diminutive} 	 & 10.0 & & 3.6     & 5.8	& & 0.0     & 3.1   & & 2.1  	& 20.7		  	& & 1.7  	& 10.1		     & & 14 			& \textbf{44.9}\\
\textbf{Plural} 		 & 3.8 	& & 44.4    & 36.2  & & 20.9    & 10.5  & & 34.9 	& 37.5		  	& & 42.2   	& 43.9		     & & \textbf{57.4} & 56.1\\
\textbf{Opposites} 		 & 0.0 	& & 7.3     & 4.0	& & 0.6     & 0.4   & & 0.5  	& 7.0			& & 2.2  	& 12.9		     & & 8.2 			& \textbf{22.1}\\
\textbf{Currency} 		 & 4.4 	& & 2.7     & 10.0  & & 2.2     & 0.0   & & 4.5  	& 7.6			& & 2.6  	& 12.1		     & & 2.7			& \textbf{15.0}\\
\textbf{Nationalities} 	 & 2.6 	& & 1.2     & 20.6  & & 0.8     & 4.0   & & 5.1  	& 14.4		  	& & 3.7  	& \textbf{21.6}  & & 3.1 		    & 21.4\\
\textbf{Country} 		 & 1.9 	& & 20.2    & 47.1  & & 2.2     & 2.8   & & 14.3 	& 36.6		  	& & 22.8 	& \textbf{52.1}  & & 25.1 		    & 43.1\\
\textbf{Gender} 		 & 25.3 & & 30.6    & 52.9  & & 25.2    & 21.9  & & 44.7 	& \textbf{75.9} & & 45.1 	& 64.9		     & & 50.7			& 72.5\\
\midrule
\textbf{Average} 		 & 6.5 	& & 19.6    & 31.0 & & 10.3 	& 16.3  & & 23.6 	& 42.0		  	& & 26.5 	& 38.1		     & & 34.7			& \textbf{51.3}\\

\bottomrule
\end{tabular}}
\caption{Relation Identification set categories, the performance of the Polygot baseline on this task, as well as that of SPPMI and SGNS trained on the listed corpora.}
\end{table*}

This task is based on the well-known relation identification dataset which was
included with the original \texttt{word2vec} toolkit\footnote{
\url{code.google.com/archive/p/word2vec/}}, and which includes
approximately 20,000 relation identification questions, each of the form: ``If
\emph{A} has a relation to {B}, which word has the same relation to {D}?''. As
such, it uses the fact that vectors are \emph{compositional}. For example, given
\texttt{man}, \texttt{woman}, and \texttt{king}, the answer to the question
should be \texttt{queen}, the relation here being `gender'. In the original set,
these questions were divided into several categories, some based on semantic
relations, e.g. `opposites' or `country capitals', and some based on syntactic
relations, e.g. `past tense'. Mirroring this, we created a similar evaluation
set for Dutch. Considering the categories used, we aimed to replicate the
original evaluation set as closely as possible, while also including some
interesting syntactic phenomena in Dutch that are not present in English, such
as the formation of diminutives. This resulted in 11 categories; 6 syntactic and
5 semantic. See Table~\ref{predtable} for an overview of the categories and an
example for each category. Subsequently, we created a set of words occurring in
all corpora---not taking into account word frequency---and retrieved applicable
tuples of words from this vocabulary which fit the categories. By only taking
words from the intersection of the vocabulary of all models, it is guaranteed
that no model is unfairly penalized, assuring that every model is able to
produce an embedding for each word in the evaluation set. After selecting
approximately 20 tuples of words for each category, the 2-permutation of each
set was taken separately, resulting in approximately 10,000 predicates.

As an evaluation measure the following procedure was performed for each set of
embeddings: \noindent For each 2-permutation of tuples in the predicate
evaluation set $(A, B), (C, D)$, where $A$, $B$, $C$, and $D$ are distinct
words, the following test was performed:

\begin{equation}
\underset{v \in V}{\arg\max}(\simx(v, A - B + D))
\end{equation}
\noindent
Where $\simx$ is the cosine similarity:

\begin{equation}\label{cossim}
\simx(w1, w2) = \frac{\overset{\rightarrow}{w1}.\overset{\rightarrow}{w2}}{\left\lVert w1\right\rVert \left\lVert w2\right\rVert}
\end{equation}
\noindent

The objective is thus to find the word $v$ in the vocabulary $V$ which maximizes
the similarity score with the vector $(A - B + D)$.

\subsection{Dialect Identification} \label{dialect}

The relationship evaluation set above is a test of the quality of different
embeddings. However, this does not prove the effectiveness of word embeddings as
a linguistic resource. To counteract this, we created a task in which we try to
detect dialectal variation in social media posts. The goal is to measure whether
a resource that is equivalent to a hand-crafted resource can be created without
any supervision. This identification of text containing dialect has been of
interest to researchers across different languages such as Spanish
\citep{gonccalves2014}, German \citep{scheffler2014}, and Arabic \citep{lin2014}.
The task, then, is to correctly map text with dialect-specific language to the
region of origin.

To test if the embeddings provide richer information regarding dialects than
hand-crafted dictionaries, performance for both approaches needs to be compared.
The amount of dialect groups for this task was determined based on the
correspondence between those in the dialect dictionaries and a social media test
set described in Section~\ref{sec:dd}, which resulted in an identification task
of at total 16 Dutch and Flemish provinces. For classification of dialect using
the embeddings, we use each word in a document to rank the dialects for that
document using two simple methods:

\paragraph{PROV} using this method, we classify social media posts as belonging
to a province by computing the similarity (as defined in Eq. \ref{cossim}) of
every word in the post with all province names, and label the post with the
province that was most similar to the highest amount of words. As such, we
assume that the province which is most similar to a given word in
$n$-dimensional space is the province to which that word belongs.

\paragraph{CO} like PROV, but including countries, i.e. `Nederland' and
`Belgi\"e' as possible targets. Hence, any words closer to either of
the country names will not be assigned a province.  This has a normalizing
effect, as words from the general Dutch vocabulary will not get assigned a
province. \newline

We tested both these methods for SPPMI and SGNS models. For the dictionary the
procedure was largely similar, but instead of distance a lookup through the
dictionaries was used.

\section{Results}

\subsection{Relation Identification}

\begin{figure*}[t]
\begin{tikzpicture}
    \begin{axis}[
        ymin = 0,
        ymax = 1.0,
        width  = 0.95*\textwidth,
        height = 4cm,
        major x tick style = transparent,
        ybar,
        bar width=5pt,
        ymajorgrids = true,
        ylabel = {Accuracy},
        symbolic x coords={ANT,O-VL,W-VL, LI, VL-BR},
        xtick = data,
        scaled y ticks = false,
    ]

        \addplot[style={black,fill=gray!20!white,mark=none}]
        coordinates {(ANT, 0.64)   (O-VL, 0.05)   (W-VL, 0.157)  (LI, 0.122) (VL-BR, 0.387)  };

        \addplot[style={black,fill=gray,mark=none}]
        coordinates {(ANT, 0.40)    (O-VL, 0.042)     (W-VL, 0.124)  (LI, 0.143) (VL-BR, 0.438)  };

        \addplot[style={black,fill=gray!50!white,mark=none}]
        coordinates {(ANT, 0.837)   (O-VL, 0.0)   (W-VL, 0.0)  (LI, 0.139) (VL-BR, 0.0)   };

       	\addplot[style={black,fill=gray!50!white,mark=none}]
        coordinates {(ANT, 0.85)   (O-VL, 0.0)   (W-VL, 0.0)  (LI, 0.153) (VL-BR, 0.0)   };

        \addplot[style={black ,fill=black,mark=none}]
        coordinates {(ANT, 0.163)   (O-VL, 0.226)   (W-VL, 0.088)   (LI, 0.216) (VL-BR, 0.065) };

    \end{axis}
\end{tikzpicture}

\begin{tikzpicture}
    \begin{axis}[
        ymin = 0,
        ymax = 1.0,
        width  = 0.95*\textwidth,
        height = 4cm,
        major x tick style = transparent,
        ybar,
        bar width=5pt,
        ymajorgrids = true,
        ylabel = {Accuracy},
        symbolic x coords={N-HO,Z-HO,ZEE,N-BR,GRO, FRI, GEL, UTR},
        xtick = data,
        scaled y ticks = false,
        legend style={at={(0.5,-0.4)},
        anchor=north,legend columns=-1},
    ]

        \addplot[style={black,fill=gray!20!white,mark=none}]
        coordinates {   (N-HO, 0.017)   (Z-HO, 0.394)   (ZEE, 0.217)   (N-BR, 0.083)   (GRO, 0.115)   (FRI, 0.073)   (GEL, 0.021) (UTR, 0.203)};

        \addplot[style={black,fill=gray,mark=none}]
        coordinates {   (N-HO, 0.021)   (Z-HO, 0.308)   (ZEE, 0.063)    (N-BR, 0.098)    (GRO, 0.159)   (FRI, 0.073)   (GEL, 0.027) (UTR, 0.250)};

        \addplot[style={black,fill=gray!50!white,mark=none}]
        coordinates {  (N-HO, 0.0)   (Z-HO, 0.0)   (ZEE, 0.33)   (N-BR, 0.0)   (GRO, 0.30)   (FRI, 0.12)   (GEL, 0.009) (UTR, 0.142) };

        \addplot[style={black,fill=gray!50!white,mark=none}]
        coordinates {  (N-HO, 0.0)   (Z-HO, 0.0)   (ZEE, 0.167)   (N-BR, 0.0)   (GRO, 0.374)   (FRI, 0.044)   (GEL, 0.007) (UTR, 0.173) };

        \addplot[style={black,fill=black,mark=none}]
        coordinates {  (N-HO, 0.112)   (Z-HO, 0.126)   (ZEE, 0.049)   (N-BR, 0.306)    (GRO, 0.032)   (FRI, 0.075)    (GEL, 0.073) (UTR, 0.061)};

        \legend{SGNS PROV, SGNS CO, SPPMI PROV, SPPMI CO, Dictionary}
    \end{axis}
\end{tikzpicture}

\caption{Accuracy scores per Flemish (top) and Dutch (bottom) province per model. Scores for the provinces of \emph{Drenthe}, \emph{Flevoland} and \emph{Overijssel} are not listed, as these were not present in the test set.}
\label{dialectfig}
\end{figure*}
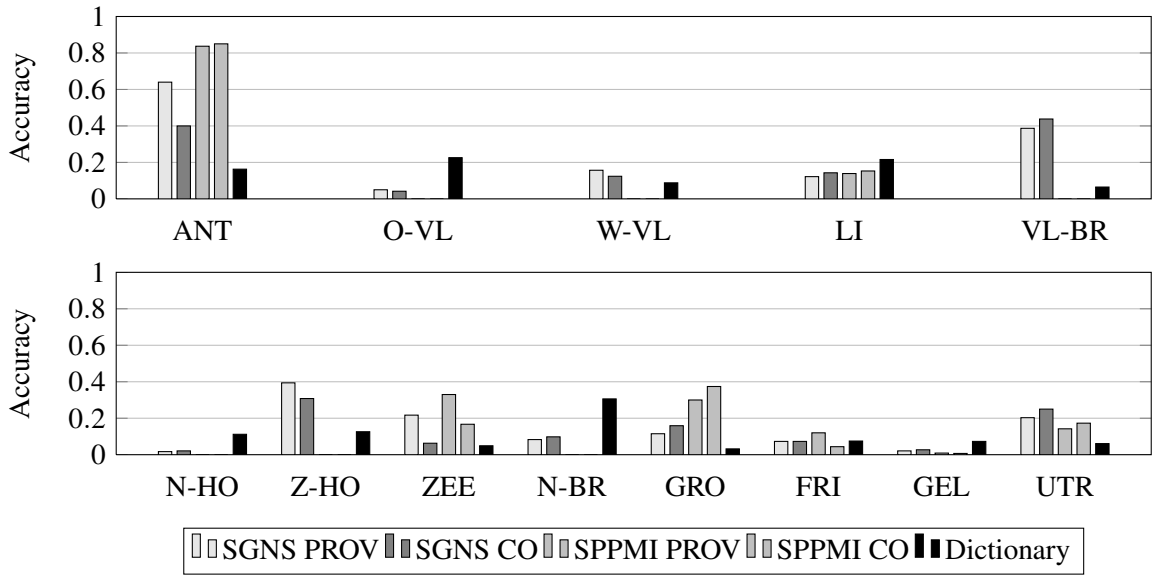

The results of the experiment on the relation identification are presented in
Table~\ref{predtable}, which shows that all models obtain higher performance on
the syntactic categories when compared to the semantic categories, the exception
being the `gender' category, on which all models did comparatively well.
Furthermore, performance on `currency' and `opposites' was consistently low, the
former of which could be explained through low occurrence of currencies in our
data. All models outperform the baseline embeddings, which is made all the more
problematic by the fact that the vocabulary of the baseline model was fairly
small; only 6000 out of the 10,000 predicates were in vocabulary for the model.
While it is not possible to estimate how the model would have performed on OOV
(Out Of Vocabulary) words, this does demonstrate that it performs well
even given a large variety of words.

\begin{table}
\centering
\small
\begin{tabular*}{\columnwidth}{@{\extracolsep{\fill}} lrrrrr}
\toprule
\multirow{2}{*}{} &
\multicolumn{2}{c}{\textbf{SGNS}} &
\multicolumn{2}{c}{\textbf{SPPMI}} &
\multicolumn{1}{c}{\textbf{DICT}} \\
& PROV & CO & PROV & CO & PROV \\
\midrule
\textbf{Acc}        & 16.4\% 		& 13.6\%   		 & 17.1\% 		   & \textbf{17.8\%} & 9.2\% \\
\textbf{MRR}        & \textbf{27\%} & 21.1\%   		 & 22.1\%		   & 22\% 			 & 14.3\% \\
\bottomrule
\end{tabular*}
\caption{Accuracy and MRR scores for SGNS, SPPMI , and the dictionary.}
\label{dialect1}
\end{table}

Comparing different SGNS models, it is safe to say that the biggest determiner
of success is corpus size: the model based on the largest corpus obtains the
highest score in 7 out of 11 categories, and is also the best scoring model
overall. The Roularta embeddings, which are based on the smallest corpus,
obtained the lowest score in 7 categories, and the lowest score overall. More
interesting is the fact that the Combined corpus, does not manage to outperform
the SoNaR corpus individually. This shows that combining corpora can cause
interference, and diminish performance.

Given the purported equivalence of SPPMI and SGNS, it is surprising that the
performance of the SPPMI models was consistently lower than the performance of
the SGNS models, although the SPPMI COW model did obtain the best performance on
plurals. None of the SPPMI models seem to capture
information about superlatives or nationalities reliably, with all scores for
superlatives close to 0, and (with the exception of the COW corpus) very low
scores for nationality.

Finally, \cite{mikolov2013} report comparable performance (51.3 average) on
the English variant of the relation dataset. While this does not reveal anything
about the relative difficulty of the predicates in the dataset, it does show
that our Dutch set yields comparable performance for a similar architecture.

\subsection{Dialect Identification}

As the models based on the COW corpus obtained the best results on the previous
task, we used these in the dialect identification task. To determine the
validity of using these models on our test data, we report coverage percentages
for the models and dictionaries with regards to the test data vocabulary. The
dialect part of our hand-crafted dictionaries had a coverage of 11.6\%, which
shows that the test set includes a large part of dialect words, as expected. The
Dutch part of the dictionary covered 23.1\% of the corpus. The SGNS model had a
coverage of 68.3\%, while the SPPMI model had a coverage of 24.4\%, which is
fairly low when compared to the SGNS model, but still more than either of the
dictionaries in separation.

As our methods provide a ranking of provinces, both accuracy and mean reciprocal
rank (MRR) were used to evaluate classification performance. While accuracy
provides us with a fair measure of how well a dialect can be predicted for a
downstream task, MRR can indicate if the correct dialect is still highly ranked.
As summarized in Table~\ref{dialect1}, SPPMI obtained the highest accuracy score
when countries were included as targets. When MRR was used as a metric, SGNS
obtained the highest performance.

Performance per dialect is shown in Figure \ref{dialectfig}. Here, SGNS
embeddings outperform the dictionaries in 7 out of 13 cases, and the SPPMI
models outperform both the SGNS and dictionary models on several provinces.
Regarding SPPMI, the figure reveals a more nuanced pattern of performance: for
both tasks, the SPPMI model obtains surprisingly high performance on the ANT
dialect, while having good performance on several other dialects. This is
offset, however, by the fact that the model attains a score of 0\% on 6
provinces, and a very low score on 2 others. An explanation for this effect is
that, being derived from a very large co-occurrence matrix, SPPMI is less able
to generalize and more prone to frequency effects. To find support for this
claim, we assessed the corpus frequencies of the province names in the COW
corpus, and found that the names of all 6 provinces on which the SPPMI models
obtained a score of 0 had a corpus frequency which was lower than 700. To
illustrate; the name of the first high-frequent province, Overijssel, for which
we do not have labeled data, has a frequency of 35218. Conversely, the provinces
of Utrecht (UTR), Groningen (GRO), and Antwerpen (ANT) are all very
high-frequent, and these are exactly the provinces on which the SPPMI model
obtains comparably high performance. While the SGNS model showed a similar
pattern of performance, it scored better on provinces whose names have a high
corpus frequency, showing that it is influenced by frequency, but still is able
to generalize beyond these frequency effects.

\section{Conclusion}

In this paper, we provided state-of-the-art word embeddings for Dutch derived
from four corpora, comparing two different algorithms. Having high
dimensionality, and being derived from large corpora, we hypothesized they were
able to serve as a helpful resource in downstream tasks. To compare the
efficiency of the embeddings and the algorithms used for deriving them, we
performed two separate tasks: first, a relation identification task, highly
similar to the relation identification task presented with the original
\texttt{word2vec} toolkit, but adapted to specific phenomena present in the
Dutch language. Here we showed to obtain better performance than the baseline
model, comparable to that of the English \texttt{word2vec} results for this
task. Secondly, a downstream dialect identification task, in which we showed
that both methods we use for deriving word embeddings outperform expensive
hand-crafted dialect resources using a simple unsupervised procedure.

\section{Acknowledgements}

We would like to thank TextGain for making the corpora of social media posts
available to us, and \'{A}kos K\'{a}d\'{a}r for helpful remarks on our work.
Part of this research was carried out in the framework of the Accumulate IWT SBO
project, funded by the government agency for Innovation by Science and
Technology (IWT).

\bibliographystyle{apalike}
\bibliography{refs}

\end{document}